%% file: 00-main.tex
\documentclass[sigconf]{acmart}

\newcommand{\etal}{{\textit{et al.}}}

\usepackage{algorithm}
\usepackage{algorithmicx}
\usepackage{amsmath}
\algnewcommand{\LeftComment}[1]{\Statex \(\triangleright\) #1}
\usepackage{threeparttable}
\usepackage{algpseudocode}
\usepackage{booktabs}
\usepackage{multirow}
\usepackage{rotating}

\AtBeginDocument{%
  }
\setcopyright{none}                
\renewcommand\footnotetextcopyrightpermission[1]{}  

\begin{document}

\title{Forecast Accuracy Is Not Trading Profit: Evolving Small Recurrent Networks for Stock Return Prediction}

\author{Jonathan Chang}
\email{jonathanbokeunchang@gmail.com}
\affiliation{%
  \institution{Union County Magnet High School}
  \city{Scotch Plains}
  \state{New Jersey}
  \country{USA}
}
\author{Zimeng Lyu}
\correspondingauthor
\authornotemark[1]
\email{zimeng.lyu@kean.edu}
\affiliation{%
  \institution{Kean University}
  \city{Union}
  \state{New Jersey}
  \country{USA}
}

\renewcommand{\shortauthors}{Chang et al.}

\begin{abstract}
Time series forecasting models are typically compared on pointwise error, which scores a prediction in isolation from the decision it is produced for, and a lower forecast error does not imply a better decision downstream.
A parallel debate asks whether modern transformer architectures forecast better than recurrent and other lightweight models.
We compare linear, fixed recurrent, transformer, and mixing based architectures against recurrent networks evolved by neuroevolutionary architecture search, evaluating each on forecast accuracy and on the net return of a daily long/short strategy.
All models are fit on a pooled panel, one network trained across the whole universe.
Across four mid-cap portfolios and three trading years, the evolved networks rank first on both forecast accuracy and net trading performance, while the second most accurate model loses money once positions are formed and costs are charged.
The advantage tracks a horizon match, since rank IC for the evolved networks rises from a one-day to a ten-day scoring horizon while every model above 300 parameters declines.
They are also the cheapest end to end: a CPU-only search of 16 minutes yields 66-weight networks that predict in 10.8~$\mu$s on a Raspberry Pi Zero, against transformer baselines of up to 817{,}153 parameters that require GPU training.

\end{abstract}



\maketitle
\pagestyle{plain}

\input{01-introduction}

\input{02-relatedwork}

\input{03-method}

\input{04-datasets}

\input{05-results}

\input{06-conclusion}

\bibliographystyle{ACM-Reference-Format}
\bibliography{99-references}


\end{document}

%% file: 01-introduction.tex
\section{Introduction}
Forecasting asset returns is a central problem in empirical asset pricing and quantitative investment management.
Portfolio allocation, risk budgeting, and systematic trading all take conditional expected returns as their primary input, so better return forecasts translate into position sizing, realized Sharpe ratio, and the robustness of a strategy to turnover and transaction costs~\cite{markowitz1952portfolio,bali2016empirical,stulz2008rethinking}.
Returns rather than prices are the usual prediction target: prices are non-stationary whereas returns are approximately stationary in their first two moments, returns are scale free and therefore comparable across the cross section, and returns map directly into the expected return and covariance structure that an allocator optimizes~\cite{campbell1997econometrics,tsay2005analysis,fama1993common}.

Stock return forecasting has moved from ARIMA and related statistical models to recurrent networks, and more recently to transformer architectures, yet which family forecasts best remains contested.
Transformers are argued to capture long range dependencies and to suit the noisy, non-stationary, multivariate series encountered in practice, while a recent line of work reports that lightweight alternatives match or exceed them on standard forecasting benchmarks~\cite{zeng2023transformers}.
That comparison is conducted almost entirely on pointwise error criteria such as MSE and MAE, which score a forecast in isolation from the decision it is produced for.
This is a general limitation of forecasting benchmarks: a small improvement in MSE or MAE does not imply that the forecast serves the downstream decision any better.
The gap is pronounced in return forecasting, where the predicted return of each stock directly drives the trading decision.

Three further design questions are left open by this literature.
The first is how a panel of multiple stocks should be arranged for training, whether to fit one model per stock or a single model across all stocks, and in what form the examples should be presented to it.
The second is whether state of the art time series models forecast well enough to support trading decisions, in the sense of outperforming a buy-and-hold allocation.
The third is whether their computational cost is warranted, since transformers and other recent architectures generally require GPUs for training and inference.

This paper investigates these questions by comparing model families for stock return forecasting in an investment portfolio setting.
The comparison covers DLinear~\cite{zeng2023transformers}, a linear model; the fixed-topology recurrent networks LSTM-i and GRU-i; the transformer architectures PatchTST~\cite{nie2022time}, iTransformer~\cite{liu2024itransformer}, and DeformTime~\cite{shu2024deformtime}; and TimeMixer~\cite{wang2024timemixer}, a decomposable multiscale mixing model built on MLPs rather than attention.
Alongside these fixed architectures we apply neuroevolutionary neural architecture search, using EXAMM~\cite{ororbia2019examm} to evolve recurrent topologies, memory cell types, and deep recurrent connections directly from the data.
Every model is fit on a pooled panel, in which one model is trained across all stocks with no stock identifier among the inputs.
Predictions are converted into positions by a daily long/short strategy charged the half quoted spread on every trade, so that forecast quality is assessed together with the decisions it produces.
Models are compared on information coefficient, net return, Sharpe ratio, and maximum drawdown across four disjoint mid-cap portfolios and three trading years, against buy-and-hold and daily equal-weight benchmarks.

This paper makes four contributions.
It introduces a pooled panel construction for neuroevolutionary architecture search on equity data, ablated against per-stock and concatenated alternatives, and shows that its benefit lies in cross-sectional ranking rather than in forecast error.
It provides a controlled comparison of neuroevolved recurrent networks against linear, fixed recurrent, and attention based families under an identical universe, walk-forward protocol, trading rule, and cost model, evaluated on both forecast accuracy and realized trading performance.
It shows that a model can rank the cross section well on average and still lose money once positions are formed and costs are charged.
It annualized cost end to end, from CPU-only architecture search to on-device inference, and finds that the smallest models in the study are also the strongest.

%% file: 02-relatedwork.tex
\section{Related Work}
Traditional models such as ARIMA have long been used for stock forecasting~\cite{reddy2019predicting}.
Machine learning models capture non-linear predictor interactions, and Gu~\etal~report that trees and networks outperform penalized linear methods on the U.S. cross section~\cite{gu2020empirical}.
Deep models differ mainly in how they encode temporal and cross-sectional structure.
Recurrent networks are the most common choice, applied to price levels~\cite{smith2024stock,ghosh2019stock} and to trend and direction~\cite{zhao2021prediction,yao2018high}.
Convolutional models read price and volume charts as images~\cite{jiang2023re}, graph models learn dependencies between stocks on a graph generated from the data rather than from predefined relations~\cite{you2024dgdnn}, and MASTER conditions attention on market state~\cite{li2024master}.
Recent work applies general purpose forecasters directly to financial panels, including selective state space models~\cite{liu2025attention} and language models over newsflow~\cite{guo2025exploring}, while diffusion models have been developed for time series forecasting more broadly, conditioning the denoising process on representations learned from history~\cite{alkilane2025expert}.
Whether the added capacity is necessary remains contested, and Zeng~\etal~show that DLinear, a single linear layer with seasonal-trend decomposition, matches or exceeds transformer forecasters on standard benchmarks~\cite{zeng2023transformers}.
Rahimikia~\etal~report that off-the-shelf foundation models transfer poorly to financial panels, and that pre-training on financial data recovers economic value~\cite{rahimikia2025re}.

In the cross-sectional setting a forecast is consumed by an allocation rule, and the daily long-short book is the standard vehicle~\cite{hou2020enriched,park2020intelligent}.
The forecasters compared in this work span the families used for that task.
DLinear is the linear baseline of the debate above~\cite{zeng2023transformers}, PatchTST segments each series into patches used as tokens and treats channels independently~\cite{nie2022time}, iTransformer tokenizes each variate so attention runs across the cross section rather than along time~\cite{liu2024itransformer}, and DeformTime uses deformable attention to follow dependencies that shift across both time and variates~\cite{shu2024deformtime}.
TimeMixer drops attention entirely and mixes decomposed components at multiple scales with MLPs~\cite{wang2024timemixer}.
Recurrent predictors also appear throughout portfolio construction~\cite{cao2020delafo,sen2021stock}, while reinforcement learning instead trains a policy on the trading objective directly~\cite{zou2024novel}.
Lee~\etal~show that a decision-focused objective tilts squared error by the inverse covariance matrix, raising prediction error while improving realized portfolio performance~\cite{lee2025return}.

Evolutionary methods improve a population of candidates through selection, mutation, and crossover~\cite{mccall2005genetic}, and in neuroevolution the search runs over network structure while the weights of each candidate are trained by gradient descent.
Early applications in finance searched around a fixed predictor, for feature selection coupled with LSTM networks~\cite{chen2020stock}, for the time window and hidden layer configuration of an LSTM~\cite{chung2018genetic}, and for information fusion in price and trend prediction~\cite{thakkar2022information}, while Jozefowicz~\etal~evolved recurrent cell structures outside finance and found variants competitive with the LSTM~\cite{jozefowicz2015empirical}.
This work builds on EXAMM, which evolves recurrent topologies, memory cell types, and deep recurrent connections from minimal seed networks~\cite{ororbia2019examm}, maintains diversity by island extinction and repopulation~\cite{lyu2021improving}, and has been extended to non-stationary streams~\cite{lyu2022one,lyu2023online}.
Lyu~\etal~apply it to equity forecasting and report evolved recurrent networks competitive with transformers at a fraction of their parameter count~\cite{lyu2025evolving}.

%% file: 03-method.tex
\section{Method}

\subsection{Panel Construction}
\label{sec:problem}

Models that are fit on many stocks at once have been shown to deliver strong and economically significant returns~\cite{gu2020empirical}.
The same holds in time series forecasting, where one global model usually beats one model per series.
Training on many related series gives the model more data to learn from, and pushes it to pick up dynamics that are shared across series instead of fitting the noise in any single one~\cite{montero2021principles}.

Let $x_{i,t}\in\mathbb{R}^{d}$ be the predictors of stock~$i\in\{1,\dots,N\}$ on trading day~$t$, and let $y_{i,t}=r_{i,t+1}$ be its next-day return.
The same panel can be arranged in three ways.
\emph{Individual} keeps the $N$ series separate and fits one model per stock, each mapping $x_{i,t}\mapsto y_{i,t}$ with $d$ inputs and one output.
\emph{Concatenated} stacks the cross-section into wide format and fits a single model
\begin{equation}
(x_{1,t},\dots,x_{N,t})\ \mapsto\ (y_{1,t},\dots,y_{N,t}),
\end{equation}
with $Nd$ inputs and $N$ outputs, so both dimensions grow linearly in~$N$.
\emph{Pooled} keeps the $d$ inputs and one output of the individual construction, but fits a single model on the union of all stocks' examples,
\begin{equation}
\mathcal{D}=\bigcup_{i=1}^{N}\{(x_{i,t},y_{i,t})\}_{t},
\end{equation}
with no stock identifier among the inputs.
Input dimensionality is therefore fixed at $d$ for any~$N$, while the number of training examples grows in proportion to~$N$.

We adopt the pooled construction and ablate the three empirically in Section~\ref{res:pooled}.

\subsection{EXAMM}
\label{subsec:examm}

This work uses the Evolutionary eXploration of Augmenting Memory Models (EXAMM) neuroevolution algorithm~\cite{ororbia2019examm} to evolve RNNs for stock return prediction. 
EXAMM searches by repeating a simple loop.
It selects one or two parents from the current population, applies a mutation or crossover operator to produce a child genome, and trains that child for a small number of epochs.
The child is inserted into the population if its validation error improves on that of the worst current member.
Runs begin from minimal seed networks containing only input-to-output connections, so a network grows only as the search finds structure worth keeping.
The 11 mutation and 2 crossover operators~\cite{ororbia2019examm} act at the level of individual nodes and edges.
They add or remove a node, add or split an edge, add a recurrent edge with a randomly selected time delay, or change the memory cell type of a node.

Two features distinguish the resulting architectures from a fixed standard RNN.
First, node type is part of the search.
EXAMM draws from a library of $\Delta$-RNN units~\cite{ororbia2017diff}, GRUs~\cite{chung2014empirical}, LSTMs~\cite{hochreiter1997long}, MGUs~\cite{zhou2016minimal}, and UGRNNs~\cite{collins2016capacity}, and may mix cell types within a single network, so the memory mechanism is selected per node from data rather than fixed a priori across the whole model.
Second, EXAMM evolves \emph{deep recurrent connections}, recurrent edges that span more than one time step.
These let a network read information from arbitrary past lags directly, without propagating it through every intermediate state, and have been shown to improve generalization and to yield models that outperform standard gated architectures~\cite{desell2019evostar-deeprecurrent}.

EXAMM is an asynchronous distributed search algorithm.
A master process maintains the population and issues genomes on request, while workers independently train genomes and report their fitness back.
No worker waits on any other, so the search scales to the number of available cores without a synchronization barrier.
Diversity is maintained structurally: the population is partitioned into islands that evolve independently and exchange genetic material only through periodic inter-island crossover, and islands that stagnate are erased and repopulated from the best genomes elsewhere in the population~\cite{lyu2021improving}.

\subsection{Daily Long--Short Trading Strategy}
\label{subsec:trading}

Predictions are evaluated through a daily long--short strategy, the standard choice in the related literature~\cite{park2020intelligent, hou2020enriched}.
The book is parameterized by $K_L$ and $K_S$, the number of long and short positions.
Short notional equals the initial capital, so \$100 of capital supports a \$200 book with half on each side.

Algorithm~\ref{alg:long-short} gives the procedure. On each trading day
$t$, stocks are sorted by predicted return. The book is rebalanced only
when the $K_L$ top-ranked stocks all have positive predicted returns and
the $K_S$ bottom-ranked all have negative; otherwise existing positions
are held. On a rebalance, all positions are liquidated and capital is
reallocated equally within each side. The gate is identical across
models, but how often it opens depends on their predictions, so holding
periods differ across the comparison.

\begin{algorithm}[!t]
    \caption{Daily Long--Short Strategy}
    \label{alg:long-short}
    \begin{algorithmic}[1]
        \footnotesize
        \Function{LongShortReturn}{$P$, $C_0$, $K_L$, $K_S$}
            \State $C \gets C_0$
            \For{$t$ in $[1,T]$}
                \State $S \gets \textsc{SortByPrediction}(P, t)$
                \If{$S[K_L].\hat{y}_t > 0$ \textbf{and} $S[-K_S].\hat{y}_t < 0$}
                \State $C \gets C + \textsc{ClearHoldings}(P, t)$ \Comment{\emph{sell longs, cover shorts}}
                \State $q_L \gets C/K_L$, \enspace $q_S \gets C/K_S$
                \For{$i$ in $[1,K_L]$}
                    \State $S[i].\textsc{Buy}(q_L, t)$, \enspace $C \gets C - q_L$
                \EndFor
                \For{$i$ in $[1,K_S]$}
                    \State $S[-i].\textsc{Short}(q_S, t)$, \enspace $C \gets C + q_S$
                \EndFor
                \EndIf
            \EndFor
            \State $C \gets C + \textsc{ClearHoldings}(P, T)$
            \State \Return $(C - C_0)/C_0$ \Comment{\emph{net return over the test period}}
        \EndFunction
    \end{algorithmic}
\end{algorithm}

%% file: 04-datasets.tex
%
%

\section{CRSP Dataset}
\label{sec:dataset}

The Center for Research in Security Prices (CRSP) dataset is the standard academic source for U.S. equity prices, returns, and trading volumes, and it is widely used for empirical asset pricing and financial forecasting research. The datasets used in this paper are derived from the CRSP dataset, using daily frequency data. 

We further processed the CRSP dataset by adopting the set of economic predictors used in Lyu~\etal~\cite{lyu2025evolving}, which were generated from the CRSP dataset. In total, we used six predictors: \textit{Return}, \textit{Volume Change}, \textit{Bid-Ask Spread}, \textit{Illiquidity}, \textit{Turnover}, and \textit{S\&P 500 Return}. We construct two families of portfolios from these predictors: a \emph{baseline portfolio} that replicates the stock selection of prior work, and four \emph{validation portfolios} used to test generalization beyond that fixed selection.

\subsection{Baseline Portfolio}
\label{subsec:baseline-portfolio}
We adopt the stock selection of Lyu et al.~\cite{lyu2025evolving}: 50 mid-cap companies drawn from the S\&P 500, chosen because mid-cap stocks receive less analyst and institutional scrutiny than large-caps and so are less efficiently priced. Retaining the same 50 tickers lets our results sit directly alongside that paper's, since no difference in performance can be attributed to a different forecasting universe. This dataset contains the same six predictors described above, obtained directly from CRSP or computed from CRSP variables.

\subsection{Validation Portfolios}
\label{subsec:validation-portfolios}

To test the robustness of EXAMM beyond the fixed 50-stock selection above, we construct four additional portfolios, $V1$--$V4$, of 50 stocks each.
The target universe $\mathcal{U}$ covers mid-cap through low-large-cap securities, defined by current market capitalization $2 \le M_i < 50$ in USD billions for security~$i$.
Restricting the S\&P~500 CRSP panel to securities with at least 20 years of continuous trading history leaves 203 names in this range, from which 200 were drawn at random to form $\mathcal{U}$.
These 200 stocks were then shuffled and split evenly into $V1$--$V4$, so the four portfolios are disjoint and no stock appears in more than one.

This construction carries survivorship bias, since drawing from current S\&P~500 constituents with 20 years of continuous history projects today's survivors backward into the training and test periods. The length requirement follows from the models under comparison: the RNN and transformer families evaluated here are both data-hungry, and the walk-forward folds train each of them on 17–19 years of daily data. The bias applies uniformly across models, since every family is trained and evaluated on the identical universe, folds, and trading rule, and the reported returns should be read as relative rather than as absolute levels.

\subsection{Pooled Panel and Walk-Forward Evaluation}
\label{sec:pooled-panel-construction}

In the pooled construction, each stock keeps its own six-predictor file, as in the \emph{individual} construction, but the 50 stocks are merged into a single training set with no stock identifier among the inputs.
One model is therefore fit across the whole universe and must learn a single rule that generalizes across stocks rather than 50 stock-specific ones.

Models are evaluated on two walk-forward folds of the Baseline Portfolio (Table~\ref{tab:folds}).
Each fold trains through a given year-end, validates on the following year, and tests on the year after: Test~2022 trains through 2020 and validates on 2021, and Test~2023 shifts the window forward by one year.
The two test years span a sharp regime shift, from the 2022 bear market to the 2023 bull recovery, so patterns learned under one regime need not transfer to the other and performing well on both folds is a harder bar than performing well on either alone.

The same design is applied to the Validation Portfolios $V1$--$V4$, with three folds each and test years 2022, 2023, and 2024 (Table~\ref{tab:validation-folds}).
This extends the evaluation by one year relative to the Baseline Portfolio and tests generalization across both stock selection and market conditions.

\begin{table}[t]
\centering
\caption{Baseline Portfolio Walk-Forward Folds}
\label{tab:folds}
\resizebox{\columnwidth}{!}{%
\begin{tabular}{l ccc ccc}
\toprule
& \multicolumn{2}{c}{Training} & \multicolumn{2}{c}{Validation} & \multicolumn{2}{c}{Test} \\
\cmidrule(lr){2-3} \cmidrule(lr){4-5} \cmidrule(lr){6-7}
Fold & Rows & Period & Rows & Period & Rows & Period \\
\midrule
Test 2022 & 3{,}290 & 2007-12-07--2020-12-31 & 252 & 2021 & 251 & 2022 \\
Test 2023 & 3{,}542 & 2007-12-07--2021-12-31 & 251 & 2022 & 250 & 2023 \\
\bottomrule
\end{tabular}%
}
\end{table}

\begin{table}[t]
\centering
\caption{Validation Portfolio Walk-Forward Folds}
\label{tab:validation-folds}

\resizebox{\columnwidth}{!}{%
\begin{threeparttable}
\begin{tabular}{l cc cc cc}
\toprule
& \multicolumn{2}{c}{Training} & \multicolumn{2}{c}{Validation} & \multicolumn{2}{c}{Test} \\
\cmidrule(lr){2-3} \cmidrule(lr){4-5} \cmidrule(lr){6-7}
Fold & Rows & Period & Rows & Period & Rows & Period \\
\midrule
Test 2022 & $4{,}038$--$4{,}252$ & 2004--2020-12-31 & 252 & 2021 & 251 & 2022 \\
Test 2023 & $4{,}290$--$4{,}504$ & 2004--2021-12-31 & 251 & 2022 & 250 & 2023 \\
Test 2024 & $4{,}541$--$4{,}755$ & 2004--2022-12-30 & 250 & 2023 & 252 & 2024 \\
\bottomrule
\end{tabular}%
\begin{tablenotes}[flushleft]
\item \textit{Note:} Row counts differ slightly across portfolios because their stocks have different numbers of available trading days.
\end{tablenotes}
\end{threeparttable}
}

\end{table}

\begin{table}[t]
\centering
\caption{Pooling ablation for EXAMM on Baseline Portfolio}
\label{tab:pooling-ablation}
\resizebox{\columnwidth}{!}{%
\begin{tabular}{l rrr rrr}
\toprule
& \multicolumn{3}{c}{Test 2022} & \multicolumn{3}{c}{Test 2023} \\
\cmidrule(lr){2-4} \cmidrule(lr){5-7}
Construction & Rank IC & MSE & Net(\%) & Rank IC & MSE & Net(\%) \\
\midrule
Individual & $+0.0001$ & $4.726\text{e-}4$ & $+10.47$ & $-0.0050$ & $3.037\text{e-}4$ & $-4.17$ \\
Concatenated & $-0.0081$ & $4.739\text{e-}4$ & $-28.48$ & $-0.0020$ & $3.113\text{e-}4$ & $-17.03$ \\
\textbf{Pooled} & $\mathbf{+0.0030}$ & $\mathbf{4.633\text{e-}4}$ & $\mathbf{+12.73}$ & $\mathbf{+0.0195}$ & $\mathbf{2.926\text{e-}4}$ & $\mathbf{+30.24}$ \\
\bottomrule
\end{tabular}%
}
\end{table}

\section{Experimental Setup}
\label{sec:experimental-setup}

\subsection{Models}
\label{subsec:models}

All models use the features and chronological split of Section~\ref{sec:dataset} and predict the next-day return of each stock. All models, including EXAMM, were evaluated using their
default hyperparameters without additional tuning.
EXAMM evolves recurrent topologies with $10$ islands of at most $10$ genomes each, an island repopulation frequency of $500$, and $10{,}000$ genomes generated per run.
Each generated genome is trained for $10$ epochs of backpropagation through time and scored by validation MSE.
New genomes are produced by mutation with probability $0.7$, intra-island crossover with probability $0.2$, and inter-island crossover with probability $0.1$.
Mutations are drawn uniformly from $10$ of EXAMM's $11$ operators, excluding \emph{split edge}.
New nodes are drawn uniformly from a library of simple neurons, $\Delta$-RNN, GRU, LSTM, MGU, and UGRNN memory cells, and recurrent connections receive a time skip sampled as $\mathcal{U}(1,10)$.

As fixed-architecture baselines we use DLinear~\cite{zeng2023transformers}, a linear model; PatchTST~\cite{nie2022time}, iTransformer~\cite{liu2024itransformer}, and DeformTime~\cite{shu2024deformtime} from the transformer family; and the mixing-based TimeMixer~\cite{wang2024timemixer}.
Each is given an input length of $96$ and an output length of $1$, the former being the most widely reported setting across these models and the latter matching the one-step-ahead target of the evolved RNNs.
All remaining hyperparameters follow the published defaults of each model. The two fixed-topology recurrent baselines, LSTM-i and GRU-i, are each a two hidden layer recurrent network whose hidden layer size equals the input size, trained for 1000 epochs of backpropagation through time at a learning rate of 1e-4. Every configuration is repeated $10$ times with independent seeds, and we report the mean over repeats.

\subsection{Trading Strategy and Costs}
\label{subsec:strategy}
Predictions are converted to positions by Algorithm~\ref{alg:long-short}: each day the cross-section is ranked by predicted return, and when the entry condition is met the strategy takes equal-weight long positions in the top $10$ stocks and equal-weight short positions in the bottom $10$.
Every trade is charged the half quoted spread $c_{i,t}=\tfrac{1}{2}(P^{\mathrm{ask}}_{i,t}-P^{\mathrm{bid}}_{i,t})$ per share, computed from the CRSP closing quotes of stock~$i$ on day~$t$, the same fields underlying the quoted spread predictor of Section~\ref{sec:dataset}.
Reported net returns subtract $c_{i,t}$ on every trade.

\begin{table*}[t]
\centering
\caption{Pearson information coefficient (IC) of Forecasting Performance across All Models}
\label{tab:ic}
\begin{threeparttable}
\footnotesize
\begin{tabular}{llrrrrrrrr}
\toprule
& &  & \multicolumn{1}{c}{Other} & \multicolumn{2}{c}{Fixed RNNs} & \multicolumn{3}{c}{Transformer-based} & \multicolumn{1}{c}{Mixing} \\
\cmidrule(lr){4-4} \cmidrule(lr){5-6} \cmidrule(lr){7-9} \cmidrule(lr){10-10}
Panel & Test yr & EXAMM & DLinear & LSTM-i & GRU-i & PatchTST & iTransf. & DeformT. & TimeMixer \\
\midrule
$V1$ & 2022 & +0.0000 & +0.0015 & \textbf{+0.0230} & \underline{+0.0229} & +0.0149 & +0.0138 & +0.0029 & -0.0079 \\
$V1$ & 2023 & \underline{+0.0145} & +0.0079 & +0.0050 & +0.0026 & +0.0076 & -0.0017 & -0.0022 & \textbf{+0.0164} \\
$V1$ & 2024 & \textbf{+0.0257} & +0.0117 & -0.0093 & -0.0138 & \underline{+0.0118} & -0.0093 & +0.0025 & +0.0041 \\
\addlinespace
$V2$ & 2022 & +0.0041 & -0.0010 & \underline{+0.0150} & \textbf{+0.0159} & +0.0047 & +0.0096 & +0.0005 & -0.0041 \\
$V2$ & 2023 & \underline{+0.0213} & +0.0083 & -0.0047 & -0.0032 & +0.0063 & -0.0042 & -0.0061 & \textbf{+0.0285} \\
$V2$ & 2024 & \textbf{+0.0181} & -0.0013 & -0.0083 & -0.0066 & -0.0077 & -0.0148 & -0.0118 & \underline{+0.0134} \\
\addlinespace
$V3$ & 2022 & +0.0115 & +0.0035 & +0.0100 & +0.0089 & +0.0043 & \textbf{+0.0270} & -0.0064 & \underline{+0.0142} \\
$V3$ & 2023 & -0.0171 & -0.0077 & -0.0088 & -0.0130 & -0.0121 & \textbf{+0.0080} & -0.0148 & \underline{-0.0063} \\
$V3$ & 2024 & \underline{+0.0128} & \textbf{+0.0141} & +0.0002 & -0.0002 & +0.0098 & +0.0038 & +0.0011 & -0.0003 \\
\addlinespace
$V4$ & 2022 & \textbf{+0.0272} & +0.0087 & \underline{+0.0257} & +0.0175 & +0.0220 & +0.0174 & +0.0044 & -0.0001 \\
$V4$ & 2023 & -0.0011 & \textbf{+0.0245} & -0.0150 & -0.0124 & \underline{+0.0129} & +0.0110 & -0.0007 & -0.0199 \\
$V4$ & 2024 & \textbf{+0.0107} & +0.0103 & +0.0092 & \underline{+0.0106} & +0.0078 & -0.0052 & +0.0080 & +0.0006 \\
\midrule
\multicolumn{2}{l}{\textit{Mean}} & \textbf{+0.0106} & +0.0067 & +0.0035 & +0.0024 & \underline{+0.0069} & +0.0046 & -0.0019 & +0.0032 \\
\bottomrule

\end{tabular}%
\begin{tablenotes}[flushleft]
\footnotesize
\item \textit{Note:} Values in \textbf{bold} are the highest IC for the trading year; \underline{underlined} values are the second highest.
\end{tablenotes}
\end{threeparttable}
\end{table*}

\begin{table*}[t]
\centering
\caption{Net return (\%) using Model's Predicted Return for Portfolio Trading}
\label{tab:net-returns}
\begin{threeparttable}
\footnotesize
\begin{tabular}{llrrrrrrrrrr}
\toprule
 & & \multicolumn{2}{c}{Benchmarks} &  & \multicolumn{1}{c}{Other} & \multicolumn{2}{c}{Fixed RNNs} & \multicolumn{3}{c}{Transformer-based} & \multicolumn{1}{c}{Mixing} \\
\cmidrule(lr){3-4} \cmidrule(lr){6-6} \cmidrule(lr){7-8} \cmidrule(lr){9-11} \cmidrule(lr){12-12}
Panel & Test yr & B\&H & EW & EXAMM & DLinear & LSTM-i & GRU-i & PatchTST & iTransf. & DeformT. & TimeMixer \\
\midrule
$V1$ & 2022 & -7.16 & -14.26 & -4.63 & +6.15 & \textbf{+7.13} & \underline{+6.65} & +3.54 & +0.80 & +2.48 & -0.18 \\
$V1$ & 2023 & +5.18 & -1.67 & \underline{+21.34} & -4.27 & -5.20 & +1.48 & -10.61 & -0.94 & -4.64 & \textbf{+23.46} \\
$V1$ & 2024 & -4.73 & -9.92 & \textbf{+17.19} & -15.48 & -15.49 & -15.57 & -23.11 & -17.28 & +1.73 & \underline{+13.39} \\
\addlinespace
$V2$ & 2022 & -11.42 & -18.91 & \textbf{+35.22} & +5.66 & -4.05 & +0.45 & -1.85 & \underline{+13.72} & -1.61 & -25.23 \\
$V2$ & 2023 & +15.37 & +7.40 & \underline{+16.92} & +3.29 & -22.25 & -16.90 & -10.35 & -10.34 & -15.06 & \textbf{+28.56} \\
$V2$ & 2024 & +12.13 & +3.46 & \textbf{+13.80} & -12.99 & -25.10 & -25.39 & -12.85 & -17.71 & -14.20 & \underline{-2.76} \\
\addlinespace
$V3$ & 2022 & -11.39 & -19.08 & \textbf{+36.68} & -1.43 & +17.38 & +17.99 & +4.27 & +19.65 & +3.88 & \underline{+27.55} \\
$V3$ & 2023 & +14.77 & +7.05 & \underline{-3.82} & \textbf{+1.03} & -10.61 & -13.03 & -8.87 & -18.84 & -9.69 & -13.16 \\
$V3$ & 2024 & +13.11 & +3.89 & -2.88 & \textbf{+24.88} & -1.82 & -2.98 & \underline{+11.46} & +6.32 & -4.84 & +5.62 \\
\addlinespace
$V4$ & 2022 & -10.25 & -17.33 & -5.43 & +8.14 & +11.12 & +11.91 & \underline{+20.51} & \textbf{+22.58} & -4.32 & +4.82 \\
$V4$ & 2023 & +9.41 & +3.16 & +1.84 & \textbf{+24.33} & -20.26 & -24.45 & -7.40 & \underline{+10.93} & -11.13 & -21.61 \\
$V4$ & 2024 & +6.69 & -1.14 & +5.96 & -6.26 & +1.69 & +5.13 & \textbf{+16.68} & -12.38 & -9.88 & \underline{+8.65} \\
\midrule
\multicolumn{2}{l}{\textit{Mean}} & +2.64 & -4.78 & \textbf{+11.02} & +2.75 & -5.62 & -4.56 & -1.55 & -0.29 & -5.61 & \underline{+4.09} \\
\bottomrule
\end{tabular}%
\begin{tablenotes}[flushleft]
\footnotesize
\item \textit{Note:} Values in \textbf{bold} are the highest net return for the trading year; \underline{underlined} values are the second highest.
\end{tablenotes}
\end{threeparttable}
\end{table*}

\subsection{Factor Benchmarks}
\label{sec:factors}

A net return does not establish that a strategy earns more than
compensation for exposures obtainable without a forecast. We regress
each model's daily strategy return on three benchmarks built from the
panel it trades. \textbf{MKT} is the equal-weight cross-sectional mean
return of the panel's own 50 stocks, the relevant opportunity set since
the strategy never trades outside it. \textbf{STR1} is a one-day
reversal portfolio, long the bottom 10 and short the top 10 names
ranked by day $t-1$ return; \textbf{STR5} is built identically on the
trailing five-day compounded return.

Both reversal portfolios use a $10/10$ book, matching
$K_L = K_S = 10$ in Algorithm~\ref{alg:long-short}, so factor and
strategy span the same portfolio space. The control is material: a
long--short portfolio sorted on \emph{next-day} predicted return is
structurally close to a reversal bet, and alpha that does not survive
STR1 and STR5 is better described as a repackaging of reversal than as
a forecasting result.

\subsection{Computing Environment}
\label{subsec:compute}

EXAMM is CPU only and MPI parallelized, and each run used $32$ cores of a shared allocation on dual-socket $64$-core AMD EPYC ``Milan'' nodes at [Removed for Review].
The baselines were trained in PyTorch on NVIDIA RTX~4090 GPUs.

%% file: 05-results.tex
\section{Results}
\label{sec:results}

\subsection{Pooled vs.\ Alternative Constructions}
\label{res:pooled}
Table~\ref{tab:pooling-ablation} compares the three panel constructions of Section~\ref{sec:problem} on the Baseline Portfolio, reporting test-period IC, MSE, and the net return of the long-short rule of Algorithm~\ref{alg:long-short}.
Pooling improves every metric in both folds, and is the construction used for all models, including the baselines, in the experiments that follow.

The MSE gain is modest, at $2.0\%$ in 2022 and $3.7\%$ in 2023, while the effect on cross-sectional ranking is larger.
Individual training gives an IC of $+0.0001$ in 2022 and $-0.0050$ in 2023, so its predictions carry little usable information about relative performance, whereas pooled training is positive in both folds.
Because the long-short rule allocates on relative rather than absolute predictions, it is this difference in ranking quality rather than the MSE gap that drives the trading result, and the 2023 fold moves from a $-4.17\%$ loss to a $+30.24\%$ gain.
Concatenated construction is worst on every metric.
Joining series end to end introduces boundary discontinuities that the recurrent state cannot distinguish from genuine transitions, whereas pooling keeps each series intact while sharing parameters across them.

\subsection{Forecast Quality}
\label{sec:examm-vs-transformers}
Transformers are strong on multivariate real-world series, though linear models have been reported to match or exceed them on standard forecasting benchmarks~\cite{zeng2023transformers}.
We compare both families against the evolved networks.

Table~\ref{tab:ic} reports the mean Pearson IC of the raw predictions, which is independent of the trading rule.
EXAMM attains the highest mean IC at $+0.0106$, ahead of PatchTST ($+0.0069$) and DLinear ($+0.0067$), and DeformTime is the only model with a negative mean ($-0.0019$).
All values are small in absolute terms, as expected for one-step-ahead daily return prediction, and no model is positive in every panel-year cell.

\subsection{Portfolio Returns across Market Regimes}
\label{sec:returns}
Table~\ref{tab:net-returns} reports net returns for each panel and trade year under the daily long-short rule, and Table~\ref{tab:net-returns-by-year} averages each column within trade years against the buy-and-hold (B\&H) and daily equal-weight (EW) benchmarks.
The two regimes covered differ sharply.
The 2022 test year is a drawdown across $V1$--$V4$, with B\&H at $-10.05\%$ and EW at $-17.39\%$, while 2023 and 2024 are recoveries with B\&H at $+11.18\%$ and $+6.80\%$.

EXAMM has the highest average return in all three trade years and the highest three-year mean at $+11.02\%$, against $+2.64\%$ for B\&H and $+4.09\%$ for TimeMixer.
It exceeds B\&H by $25.51$ points in the drawdown year and by $1.72$ points in the 2024 recovery, and trails it by $2.11$ points in 2023.
Only three of the eight models beat B\&H on the mean, and EXAMM and TimeMixer are the only two positive in all three trade years.
The per-panel results show that this average is not driven by single cells: EXAMM is negative in $4$ of $12$ panel-years, but those losses lie in $[-5.43\%, -2.88\%]$ while its four largest gains all exceed $+17\%$.
For the transformer baselines, per-panel gains of comparable size are offset by losses of similar magnitude, leaving year averages near or below zero.
We therefore read EXAMM's advantage as consistency across regimes rather than as outperformance in any one of them.

The two fixed-topology recurrent baselines return $-5.62\%$ and $-4.56\%$, among the weakest columns in the study.
This places the gain with the evolved topology rather than with recurrence and gated memory alone, since both baselines draw on the same cell types available to the search.
\subsection{Risk-Adjusted Performance}
\label{sec:alpha}
\begin{figure}[t]
\caption{Annualized Net Alpha with 95\% Confidence Intervals}
\centering
\includegraphics[width=0.6\columnwidth]{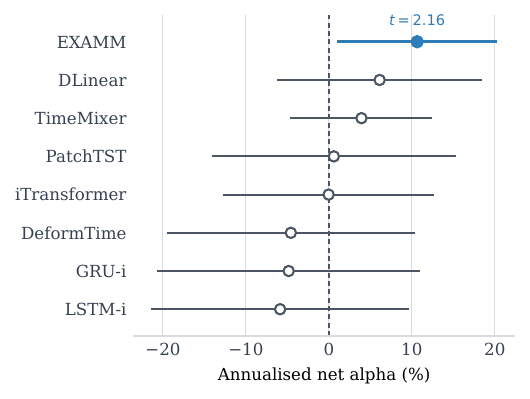}
\Description{Alpha forest plot showing annualized net alpha with 95\% confidence intervals. Only EXAMM's interval excludes zero.}
\label{fig:alpha}
\end{figure}

Figure~\ref{fig:alpha} plots the intercept of a time-series regression
of each model's daily \emph{net} strategy return on the benchmark
portfolios of Section~\ref{sec:factors}, annualized, with 95\%
confidence intervals. Regressions are estimated on the stacked
evaluation panel of $V_1$--$V_4$, comprising 23{,}424 panel-days over
732 trading dates after the STR5 warm-up. Standard errors are
Driscoll--Kraay, robust both to serial correlation within a portfolio's
return series and to contemporaneous correlation across the four
portfolios. Because the estimate is formed on the stacked panel, it
describes an equal-weight combination of the four evaluation portfolios
rather than any single 50-stock book.
 
EXAMM is the only model whose alpha is distinguishable from zero
($+10.64\%$ per year, $t = 2.16$, $p = 0.031$). Every other interval
contains zero, including those of DLinear and TimeMixer, the two models
that exceeded buy-and-hold on mean net return in
Table~\ref{tab:net-returns}. A difference in significance is not a test of difference: EXAMM's alpha
is nonzero, not demonstrably larger than DLinear's or TimeMixer's.

Table~\ref{tab:alpha-spec} reports sensitivity to the controls. Adding
one-day and five-day reversal moves alpha from $+11.27\%$ to
$+10.64\%$, an attenuation of six percent, with significance retained
throughout: the edge is not a restatement of short-term reversal,
despite the mechanical similarity between a next-day-return sort and a
reversal sort. The estimate is nonetheless imprecise, with a $95\%$
interval of $[1.0\%,\ 20.3\%]$ under the fullest specification. These
results support the claim that risk-adjusted performance is positive
and survives both reversal controls and transaction costs, not that the
strategy earns a particular premium.

Re-estimating on gross returns isolates each model's trading cost.
Under MKT, EXAMM's alpha falls from $+11.71\%$ gross to $+11.27\%$ net,
a drag of $0.44$ points per year, against $4.51$ and $4.58$ for GRU-i
and LSTM-i. Since the book turns over only when the entry condition is
met, this measures how often each model produces a cleanly signed
cross-section: the evolved networks meet it least often, hold positions
longest, and are correspondingly cheapest to trade.

\begin{table}[t]
\centering
\caption{EXAMM Annualized Net Alpha}
\label{tab:alpha-spec}
\begin{tabular}{lrrr}
\toprule
Specification & $\alpha$ (\%/yr) & $t$ & $p$ \\
\midrule
MKT                & $+11.27$ & $+2.32$ & $0.021$ \\
MKT + STR1         & $+10.96$ & $+2.24$ & $0.025$ \\
MKT + STR1 + STR5  & $+10.64$ & $+2.16$ & $0.031$ \\
\bottomrule
\end{tabular}
\end{table}

\subsection{Overall Comparison}
Table~\ref{tab:model-summary} averages four evaluation metrics over the validation portfolios, alongside each model's parameter count: information coefficient (IC), net return, Sharpe ratio, and maximum drawdown (MDD), the largest peak-to-trough decline in portfolio value. The benchmarks are listed with net return only, as buy-and-hold and equal weight issue no predictions and therefore have no IC, and their fixed allocations are not comparable on the risk metrics. EXAMM is the only model that is best on all four: IC ($+0.0106$ against $+0.0069$ for PatchTST), net return ($+11.02\%$ against $+4.09\%$ for TimeMixer), Sharpe ($+0.74$ against $+0.43$), and MDD ($-10.76\%$ against $-12.49\%$ for DLinear). Only three models return more than buy-and-hold's $+2.64\%$, and only EXAMM does so by a wide margin.

\begin{figure}[t]
\caption{Rank IC as a function of \emph{scoring} horizon}

\centering
\includegraphics[width=0.7\columnwidth]{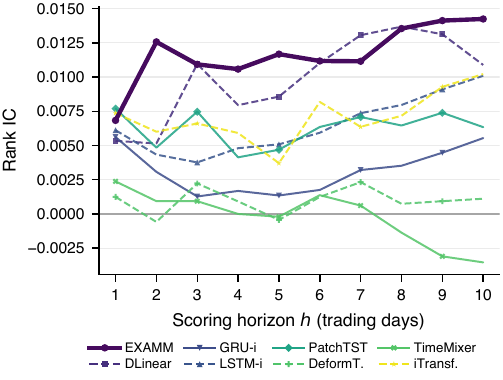}
\Description{Plot showing Rank IC as a function of scoring horizon.}
\label{fig:horizon}
\end{figure}

The parameter column shows that this does not come from capacity. EXAMM's evolved networks average $66$ weights, fewer than any other model in the comparison and two to four orders of magnitude below the Transformers ($16{,}836$ to $817{,}153$). Across the field, size and performance are unrelated or inversely related: iTransformer is the largest model at $817{,}153$ parameters and returns $-0.29\%$, DeformTime at $67{,}449$ parameters records the only negative IC in the table ($-0.0019$), while the two smallest models, EXAMM and DLinear, place in the top three on every metric. The table also separates prediction quality from tradeable profit: PatchTST has the second-highest IC ($+0.0069$) yet a net return of $-1.55\%$, so a model can rank the cross-section correctly on average and still lose money once positions are formed. EXAMM is the one model where accuracy, return, risk-adjusted return, and downside protection align, and it does so at the smallest model size. 
Figure~\ref{fig:horizon} shows where this comes from. EXAMM's rank IC rises from $+0.0068$
at a one-day horizon to $+0.0142$ at ten days, while every model above $300$ parameters
declines over the same range. Because the entry rule re-enters the book only when every
long-side prediction is positive and every short-side prediction is negative, EXAMM's
positions are held for several days rather than one, so the horizon at which its forecasts
are most accurate is also the horizon over which they are traded.

\begin{table}[t]
\centering
\caption{Average Net return (\%) across $V1$--$V4$}
\label{tab:net-returns-by-year}
\resizebox{\columnwidth}{!}{%
\begin{tabular}{lrrrrrrrrrr}
\toprule
& \multicolumn{2}{c}{Benchmarks} &  & \multicolumn{1}{c}{Other} & \multicolumn{2}{c}{Fixed RNNs} & \multicolumn{3}{c}{Transformer-based} & \multicolumn{1}{c}{Mixing} \\
\cmidrule(lr){2-3} \cmidrule(lr){5-5} \cmidrule(lr){6-7} \cmidrule(lr){8-10} \cmidrule(lr){11-11}
Test yr & B\&H & EW & EXAMM & DLinear & LSTM-i & GRU-i & PatchTST & iTransf. & DeformT. & TimeMixer \\
\midrule
2022 & -10.05 & -17.39 & \textbf{+15.46} & +4.63 & +7.89 & +9.25 & +6.62 & \underline{+14.19} & +0.11 & +1.74 \\
2023 & +11.18 & +3.99 & \textbf{+9.07} & \underline{+6.09} & -14.58 & -13.22 & -9.31 & -4.80 & -10.13 & +4.31 \\
2024 & +6.80 & -0.93 & \textbf{+8.52} & -2.46 & -10.18 & -9.70 & -1.95 & -10.26 & -6.80 & \underline{+6.23} \\
\midrule
\textit{Mean} & +2.64 & -4.78 & \textbf{+11.02} & +2.75 & -5.62 & -4.56 & -1.55 & -0.29 & -5.61 & \underline{+4.09} \\
\bottomrule
\end{tabular}%
}
\end{table}

\begin{table}[t]
\centering
\caption{Average Model Summary over Validation Portfolios.}
\label{tab:model-summary}
\resizebox{\columnwidth}{!}{%
\begin{tabular}{llrrrrr}
\toprule
 & Model & Params & IC & Net (\%) & Sharpe & MDD (\%) \\
\midrule
 & EXAMM & $66$ & \textbf{+0.0106} & \textbf{+11.02} & \textbf{+0.74} & \textbf{-10.76} \\
\midrule
Other & DLinear & $194$ & +0.0067 & +2.75 & +0.42 & \underline{-12.49} \\
\midrule
\multirow{2}{*}{Fixed RNNs}
 & LSTM-i & $679$ & +0.0035 & -5.62 & -0.15 & -18.93 \\
 & GRU-i & $511$ & +0.0024 & -4.56 & -0.10 & -19.39 \\
\midrule
\multirow{3}{*}{Transformer-based}
 & PatchTST & $16{,}836$ & \underline{+0.0069} & -1.55 & +0.12 & -15.48 \\
 & iTransf. & $817{,}153$ & +0.0046 & -0.29 & +0.17 & -15.15 \\
 & DeformT. & $67{,}449$ & -0.0019 & -5.61 & -0.19 & -17.38 \\
\midrule
Mixing & TimeMixer & $72{,}997$ & +0.0032 & \underline{+4.09} & \underline{+0.43} & -13.12 \\
\midrule
\multirow{2}{*}{Benchmarks}
 & Buy \& hold & --- & --- & +2.64 & --- & --- \\
 & Equal weight & --- & --- & -4.78 & --- & --- \\
\bottomrule
\end{tabular}%
}
\end{table}

\subsection{Scalability and Deployment}

\begin{table}[t]
\centering
\caption{EXAMM cost end to end: architecture search on CPU, and on-device inference}
\label{tab:examm-cost}
\begin{threeparttable}
\begin{tabular}{rrrr}
\toprule
\multicolumn{4}{l}{\textit{Architecture search (CPU only, $10{,}000$ genomes)}} \\
\midrule
Ranks & Workers & Wall-clock (min) & Core-h \\
\midrule
$8$   & $7$   & $72.8 \pm 15.6$ & $9.70$ \\
$16$  & $15$  & $37.1 \pm 7.7$  & $9.88$ \\
\textbf{32} & \textbf{31} & $\mathbf{16.2 \pm 2.8}$ & $\mathbf{8.65}$ \\
$64$  & $63$  & $8.1 \pm 1.9$   & $8.59$ \\
$128$ & $127$ & $\mathbf{3.9} \pm 1.0$ & $8.26$ \\
\midrule
\multicolumn{4}{l}{\textit{Deployment (Raspberry Pi Zero)}} \\
\midrule

\multicolumn{3}{l}{Parameters}          & $66$ \\
\multicolumn{3}{l}{Latency ($\mu$s)}    & $10.82$ \\
\multicolumn{3}{l}{Throughput (k/s)}    & $100.5$ \\
\multicolumn{3}{l}{Power over idle (mW)} & $476.7$ \\
\bottomrule
\end{tabular}
\begin{tablenotes}[flushleft]
\footnotesize
\item \textit{Note:} Search runs on one exclusive node, $\pm$ is the standard deviation; values in \textbf{bold} show the configuration used in all other experiments. Inference is measured on-device with an INA219 sensor.
\end{tablenotes}
\end{threeparttable}
\end{table}

Table~\ref{tab:examm-cost} reports the cost of EXAMM end to end.
The search is asynchronous: a master process maintains the population and distributes candidate genomes to workers, which train and evaluate them independently without global synchronization barriers, so scaling is near-linear in the number of workers.
Total compute stays within $8.3$--$9.9$ core-hours across all configurations tested, while wall-clock time falls from $72.8$ minutes at $8$ ranks to $16.2$ minutes at $32$, the configuration used in all other experiments, and $3.9$ minutes at $128$.
The search is CPU-only at every stage, so a complete $10{,}000$-genome run finishes in $72.8$ and $37.1$ minutes at $8$ and $16$ ranks, core counts typical of contemporary laptop and desktop processors, whereas the transformer baselines require GPU training for models of up to $817{,}153$ parameters.

The resulting models are correspondingly inexpensive to deploy.
Evolved genomes average $66$ parameters and execute in $10.82\,\mu$s per prediction on a Raspberry~Pi~Zero, a passively cooled single-core device, at $476.7$~mW above idle draw.
Inference latency is negligible relative to the daily rebalancing interval, and the sub-watt requirement permits operation on battery or embedded hardware.
The same networks that produce the returns in Table~\ref{tab:net-returns-by-year} can therefore be both discovered and deployed on commodity hardware, which is not feasible for the baseline architectures.

%% file: 06-conclusion.tex
\section{Conclusion}

This paper studied pooled panel construction for neuroevolved recurrent architecture search, and evaluated the resulting models as return forecasters for daily long-short portfolio formation.
Pooling acts on cross-sectional ranking rather than on pointwise accuracy.
Relative to per-stock training it lowers MSE by less than 4\%, but moves the information coefficient from near zero or negative to positive in both folds, and the information coefficient is the quantity a rank-based allocation consumes.
Under this construction EXAMM ranks first on information coefficient, net return, Sharpe ratio, and maximum drawdown across four disjoint portfolios and three trading years, and it remains profitable in the 2022 drawdown as well as the 2023 and 2024 recoveries. It is also the only model whose alpha against the traded universe is positively
distinguishable from zero, and that alpha survives controls for
short-term reversal, to which a next-day-return sort is mechanically
related.
The evolved networks average 66~weights, are searched on CPU in under twenty minutes on 32~cores, and evaluate in $10.82~\mu$s on a Raspberry~Pi~Zero.
Performance therefore does not follow capacity: across the eight models compared, parameter count is unrelated to realised return, and the two smallest models place in the top three on every metric.